\documentclass[man, 12pt, figsintext]{apa7}
\usepackage{newtxtext,newtxmath}
\usepackage[style=numeric, sorting=none, backend=biber, natbib=true]{biblatex}
\usepackage{graphicx}      % For images
\usepackage{booktabs}      % For professional looking tables
\usepackage{caption}       % Better captions
\usepackage{subcaption}    % For sub-figures (Grad-CAM maps)
\usepackage{float}         % To force image placement
\usepackage{amsmath}       % For math formatting
\usepackage{csquotes}      % Context sensitive quotes
\usepackage{hyperref}      % For clickable links
\usepackage{amssymb}
\nocite{*}
\usepackage{booktabs}
\usepackage[table]{xcolor}
\definecolor{HeaderBlue}{HTML}{1F4E79}
\usepackage{caption}
\usepackage{enumitem}
\usepackage[flushleft]{threeparttable}
\title{Lost in Translation: How Language Re-Aligns Vision for Cross-Species Pathology}
\shorttitle{Language Re-Aligns Vision} % Appears in the top-left header
\author{Ekansh Arora}
\affiliation{Thomas Jefferson High School for Science and Technology}

\abstract{  Foundation models are increasingly applied to computational pathology (CPath), yet their behavior under cross-cancer and cross-species transfer remains unspecified. This study investigated how fine-tuning a pathology foundation model affects cancer detection performance under same-cancer, cross-cancer, and cross-species conditions. We utilized the state-of-the-art foundation model, CPath-CLIP, to train and evaluate whole-slide image patches under zero-shot and few-shot fine-tuning settings for canine and human histopathology patches. Performance was measured using area under the receiver operating characteristic curve (AUC) across detection tasks. Few-shot fine-tuning improved same-cancer (64.9\% to 72.6\% AUC) and cross-cancer performance (56.84\% to 66.31\% AUC). Cross-species evaluation revealed that while tissue matching (e.g., Human Breast to Canine Breast) enables meaningful transfer, performance remains significantly below state-of-the-art benchmarks (H-optimus-0: 84.97\% AUC). This indicates that while the visual features are transferable, the standard vision-language alignment is suboptimal for cross-species generalization. Embedding space analysis revealed extremely high cosine similarity (>0.99) between tumor and normal prototypes. Grad-CAM shows prototype-based models remain domain-locked, while language-guided models attend to conserved tumor morphology.  To address this, we introduce 'Semantic Anchoring,' a method that uses language to provide a stable coordinate system for visual features. While we initially utilized Qwen2-1.5B for its medical expressivity, our ablation studies reveal that the primary benefit stems from the text-alignment mechanism itself, regardless of the text encoder's complexity.  Furthermore, benchmarking against H-optimus-0 reveals that CPath-CLIP's failure stems from intrinsic embedding collapse, which text alignment effectively circumvents. Smaller but consistent gains were also observed in same-cancer and cross-cancer classification, with an 8.52\% and a 5.67\% increase, respectively. We identified a previously uncharacterized failure mode in pathology foundation models: semantic collapse driven by species-dominated alignment rather than missing visual information. These results demonstrate that language acts as a control mechanism, enabling semantic re-interpretation without retraining. 
}


\begin{document}

\maketitle

\section{Introduction}

Foundation models are transforming computational pathology by enabling the analysis of large-scale histopathology images, such as gigapixel-resolution Whole Slide Images (WSIs), with minimal task-specific supervision. By learning rich visual and semantic representations from massive datasets, these models recognize patterns across diverse tissue types. Vision models learn primarily from conserved morphological structures like nuclei and glands. While organ sites differ, pathology shares a consistent visual vocabulary across tissues. This consistency enables effective cross-species training and transfer learning for vision-based models. However, as domain shifts increase, the alignment between these visual features and their semantic interpretations often becomes a bottleneck for generalization.

The behavior of foundation models under cross-domain conditions, such as different cancer types or species, remains largely unexplored. This study is motivated by the significant biological overlap between canine and human cancers, which share histological appearance, tumor genetics, molecular targets, biological behavior, and response to conventional therapies. Such commonalities suggest that human-trained foundation models should be capable of detecting cancerous features in canine tissue, making cross-species generalization a scientifically valid pursuit for advancing both human and animal health [1]. 

However, the successful deployment of these models hinges on the quality of their latent representations. A critical, yet often overlooked, failure mode is 'embedding collapse,' where a model fails to maintain separation between distinct classes (e.g., tumor vs. normal) when applied to unseen domains. If the underlying representation collapses, the model will fail to generalize regardless of the biological similarities between species. Therefore, it is essential to determine whether cross-species failures arise from genuine histological differences or from intrinsic limitations in the foundation model’s embedding space. Distinguishing between these two causes is critical: biological gaps require more data, whereas representational collapse requires improved alignment strategies to recover latent signal. 

Rather than assuming cross-species failure stems from missing visual features, we hypothesized that the semantic interface of vision-language models actively suppresses tumor-relevant signals when species shifts occur. This led us to ask: can we recover generalization without retraining, simply by re-aligning how the model interprets its own frozen features? 

Currently, the behavior of these foundation models under multiple domain shifts remains poorly characterized, with no study having jointly varied cancer type, species, and supervision level. For this reason, many clinicians do not possess guidance on when fine-tuning helps versus when it can degrade performance.  

In this study, we investigated whether the fine-tuning of these foundation models, such as CPath-CLIP, a vision model pre-trained on a vast amount of different benign and cancerous human WSIs, can affect its performance in same-cancer, cross-cancer, and cross-species detection tasks. Moreover, finding out under what conditions does fine-tuning improve or hurt model generalization will lead to the examination of the foundation model performance: when zero-shot foundation models outperform fine-tuned models when domain shifts are significant. To evaluate these questions, we systematically varied cancer types, species, and supervision levels to evaluate model performance under controlled domain shifts.  

Recent advances in large language models offer stronger semantic priors, which may improve cross-species zero-shot transfer without additional labeled data. In this work, we also integrated the Qwen2-1.5B text encoder to investigate whether improved semantic alignment enables generalization across species while preserving the main evaluation of same-cancer and cross-cancer performance. 

\subsection{Model Architecture}
We utilized CPath-CLIP, a ViT-L-14 based foundation model, with an input resolution of 336x336 pixels, pre-trained on human histopathology. The model produces a high-dimensional latent representation ($\mathbb{R}^{3328}$), which captures complex morphological features [2]. To preserve the “visual vocabulary” learned during its large-scale pre-training on human tissue, the ViT backbone was kept frozen in evaluation mode throughout experiments. We appended a trainable linear classification head to translate the frozen embeddings into binary logits for cancer detection. We also tested bottleneck adapters (r=8–256) as a middle ground with selective parameter adaptation between zero-shot and full fine-tuning. 

	To isolate the role of semantic alignment in the frozen visual embedding space, we conducted a targeted ablation replacing the CLIP text tower with Qwen2-1.5B. When fine-tuning, a 3-layer MLP projection head (1536→2048→3328) was trained using InfoNCE contrastive loss while the vision backbone remained fully frozen. This design explicitly tested whether stronger semantic priors in the language encoder alone, without any canine supervision or visual adaptation, could recover cross-species generalization. All experiments were rerun under this modified alignment. 
    
\subsection{Control Experiments}
To understand the baseline performance gap between our models, we must distinguish between their training objectives. CPath-CLIP uses vision-language contrastive learning, maximizing similarity between images and their pathology report descriptions. While this provides high-level semantic awareness, it can lead to "semantic dominance" where domain-level features (like species) drown out morphological signals. In contrast, H-optimus-0 (DINOv2) uses self-supervised distillation without text.  

To strictly validate the source of performance gains, we introduced three control experiments. First, we performed an MLP Head Ablation, replacing the Qwen LLM with a standard CLIP text encoder to test if performance stems from semantic complexity or simply the alignment mechanism. Second, we established a Baseline Quality Benchmark by evaluating H-optimus-0 on the same patch set to determine if performance issues were intrinsic to the CPath-CLIP architecture. Third, we conducted Prompt Robustness Testing to measure sensitivity to phrasing, testing species-specific, generic, histological, and ontology-guided prompts. Prompt embeddings were generated using the CLIP text encoder, and inter-class cosine similarity was computed to quantify semantic separation. 

\subsection{Datasets}
To evaluate the cross-species and cross-cancer generalization of CPath-CLIP, we utilized three distinct histopathology datasets, which represented the varying degrees of morphological and biological divergence.  

First, our primary canine dataset for breast cancer consists of 21 Whole Slide Images (WSIs) of canine mammary carcinoma, as described by Bertram \& Aubreville et al. (2020). These slides represent spontaneous tumors in dogs that closely mimic the histological patterns and clinical progression of human breast cancer, providing a robust basis for evaluation of cross-species AI performance. From these 21 slides, we extracted 22,239 patches at 40x magnification. To ensure a rigorous evaluation of few-shot learning, we utilized one slide for training (which yielded 2,048 patches). The remaining slides (20,191 patches) served as the test set. This slide-level split prevented data leakage between training and testing. 

To test cross-cancer generalization, we utilized the MITOS\_WSI\_CCMCT dataset. Mast cell tumors are biologically distinct from carcinomas, allowing for a strong base for the assessment of CPath-CLIP's cross-cancer accuracy. This dataset provided 32 annotated slides, from which we used a subset of 7 slides totaling 5,530 patches (3,189 tumor [58\%], 2,341 normal [42\%]). We included this dataset to test out-of-distribution recognition of the visual vocabulary in different malignancy.  

For testing reverse transfer and cross-species evaluation, we used the TCGA-BRCA dataset for breast cancer. We sampled 505 patches from this dataset. This dataset was used to evaluate prototype sourcing, linear probe training, and reverse-transfer stress tests between human and canine tissue. Table 1 summarizes all these splits. 

\subsection{Preprocessing Pipeline}
To ensure compatibility with the CPath-CLIP architecture and to minimize variance, we implemented a standard preprocessing pipeline like the one used to train CPath-CLIP. For each WSI, we first performed tissue-background separation using Otsu thresholding to generate a binary mask from a thumbnail. We then extracted patches of 1024 by 1024 pixels at 40x magnification for the identified tissue regions. To balance dataset representation with computational efficiency, we utilized a stride of 2048 to ensure a diverse sampling of the slide. Since histopathology slides are subject to color variation due to differences in staining procedures and hardware, we applied a Macenko stain normalizer. This technique transforms pixel values into Optical Density space, where Singular Value Decomposition is used to estimate hematoxylin and eosin (H\&E) stain vectors and maps them to a consistent reference space. This prevented the model from over-relying on color-specific artifacts. To meet the input requirements of the ViT-L/14 backbone, all 1024 by 1024 patches were resized to 336 by 336 pixels using anti-aliased bilinear interpolation, maintaining the high-fidelity nuclear and glandular details learned by the foundation model during its pre-training on human tissue.  

\subsection{Experimental Design}

To systematically assess how supervision impacts generalization, we designed a controlled experimental framework that varied cancer type, species, and level of supervision while holding the frozen CPath-CLIP backbone (Section 2.1) and preprocessing pipeline constant.  

We evaluated three fundamental learning modalities across varying data regimes: 
\begin{description}

\item[Zero-shot Baseline:] No task-specific training was performed. Predictions were generated directly from the frozen foundation model using prototype-based classification. Tumor and normal prototypes were computed as the mean embeddings of labeled patches from a held-out training slide, establishing a performance reference for the model without any adaptation. 

\item[Linear Probing (Human-to-Dog Transfer):] A trainable linear classification head was optimized using human breast cancer patches from the TCGA-BRCA dataset (505 patches) and subsequently tested on canine breast cancer. This experiment served to determine whether cross-species supervision improves performance or induces domain-specific bias. 

\item[Adapter-based Fine-tuning (Dog Few-Shot Adaptation):] To evaluate target-domain supervision, bottleneck adapters (r=8–256) and linear heads were trained on progressively larger fractions of canine mammary carcinoma patches (1\%, 3\%, 5\%, 10\%, 15\%, and 20\% of the 2,048 available training patches). Patches were sampled from a single training slide to prevent patient-level data leakage. This setup allowed us to quantify the trade-off between in-domain performance and out-of-domain generalization as supervision increases. 

\item[Training Specifications:] For all supervised experiments, models were trained for 50 epochs using binary cross-entropy loss and the Adam optimizer ($learning\ rate = 10^{-3}$, $weight\ decay = 10^{-4}$).  

\item[Semantic Anchoring (Text-Guided Zero-Shot):] To test the effect of language alignment, we replaced the default classification head with a text-based projection. We utilized both Qwen2-1.5B and the standard CLIP text encoder to generate text embeddings for 'Tumor' and 'Normal,' computing similarity scores against the frozen visual features. 

\end{description}
All results were stable across five random seeds (std. dev. < 0.5\% AUC) and not driven by stochastic variation. 

\subsection{Evaluation Metrics}
To systematically evaluate CPath-CLIP under different domain shifts, we measured performance at both the patch and slide levels, using metrics that are widely used and recognized in computational pathology. These metrics allowed us to quantify not only raw classification performance but also clinical relevance.  

Our primary metric was the Area Under the Receiver Operating Characteristic Curve (AUC-ROC). This metric provides a threshold-independent measure of model discriminability between tumor and normal tissue. A higher AUC indicates better separation of cancerous and non-cancerous patches or slides. All AUC values were computed at the patch level unless otherwise specified. In addition to AUC-ROC, we evaluated standard classification metrics to provide a more granular understanding of model behavior, such as:  

\begin{description}

\item[Precision (Positive Predictive Value):] Fraction of predicted tumor patches/slides that are actually tumors. 

\item[Recall (Sensitivity):] Fraction of true tumor patches/slides correctly identified by the model. 

\item[F1-Score:] Harmonic mean of precision and recall, useful when there is an imbalance between tumor and normal patches. 

\end{description}

As said before, all AUC values were evaluate at the patch level, where CPath-CLIP operates. However, clinical decisions are made at the slide level. We needed multiple aggregation strategies to convert these patch predictions into slide-level classifications:  

\begin{description}
    \item[Mean aggregation:] Average of all patch predictions per slide.
    \item[Max aggregation:] Maximum prediction probability across all patches in a slide.
    \item[Top-5\% aggregation: ] Average of the top 5\% of predicted tumor patches per slide.
\end{description}

\noindent\textbf{Rationale:} Different aggregation strategies capture different clinical priorities. For instance, max or top-5\% aggregation emphasizes the detection of small but high-confidence tumor regions, while mean aggregation provides robustness against outlier patches.

Performance at the slide level was reported using the same metrics (AUC, accuracy, precision, recall, F1-score), allowing a direct comparison between patch-level and clinically relevant slide-level evaluation. 

All metrics were computed separately for same-cancer performance, cross-cancer performance, and cross-species performance. This framework allowed us to quantify how fine-tuning affects in-domain performance, how fine-tuning impacts out-of-domain generalization across cancer types and species, and the trade-offs between different aggregation strategies for translating patch-level predictions to slide-level clinical decisions. By combining threshold-independent (AUC) and threshold-dependent (accuracy, precision, recall, F1) metrics, we ensured that our evaluation was robust, reproducible, and clinically interpretable.  

\subsection{Semantic Anchoring Mechanism}

\textit{Semantic Anchoring} maps frozen visual embeddings $v \in \mathbb{R}^d$ onto a task-defined semantic axis defined by text embeddings $t_c$ for each class $c$. Given a normalized visual embedding $\hat{v}$ and normalized text embeddings $\hat{t}_c$, classification is performed via cosine similarity:
$$s_c = \hat{v}^\top \hat{t}_c$$
This replaces prototype-based similarity $\hat{v}^\top \hat{p}_c$, where prototype collapse was observed. No visual parameters are updated. Performance gains, therefore, reflect a reinterpretation of existing visual features rather than the acquisition of new visual information. We identify a previously uncharacterized species-dominated semantic degeneracy in pathology vision–language models and show that it can be reversed without retraining the visual backbone.

\section{Results} 

\subsection{Dataset Summary and Experimental Scope}

As shown in Table 1, we evaluated cross-domain transfer performance using three pathology datasets spanning different cancer types and species. The primary dataset consists of canine breast carcinoma patches and was used for both training and same-cancer evaluation. Cross-cancer evaluation was performed on canine mast cell tumor patches, while cross-species evaluation was conducted on a human breast cancer dataset. 

\begin{table}[ht]
\centering
\caption{Dataset Composition Across Cancer Types and Species}
\label{tab:dataset_composition}
\begin{tabular}{lccc}
\toprule
\textcolor{HeaderBlue}{\textbf{Dataset}} &
\textcolor{HeaderBlue}{\textbf{Total Patches}} &
\textcolor{HeaderBlue}{\textbf{Tumor}} &
\textcolor{HeaderBlue}{\textbf{Normal}} \\
\midrule
Canine Breast Carcinomas & 22{,}239 & 13{,}860 & 8{,}379 \\
MITOS\_WSI\_CCMCT & 5{,}530 & 3{,}189 & 2{,}341 \\
TCGA-BRCA & 505 & 505 & 0 \\
\bottomrule
\end{tabular}
\end{table}

	Cross-cancer and cross-species datasets were only used for evaluation and were never seen during training, as shown by Table 2. 

    \begin{table}[ht]
\centering
\caption{Train--Test Split for Canine Breast Carcinoma}
\label{tab:train_test_split}
\begin{tabular}{lc}
\toprule
\textcolor{HeaderBlue}{\textbf{Split}} &
\textcolor{HeaderBlue}{\textbf{Number of Patches}} \\
\midrule
Training & 2{,}048 \\
Testing & 20{,}191 \\
\bottomrule
\end{tabular}
\end{table}

\subsection{Overall Performance Across Transfer Settings}

We first compared model performance across same-cancer, cross-cancer, and cross-species transfer settings using AUC-ROC, as Table 3 shows. Notably, generic text anchoring reduced performance in the cross-cancer setting, likely due to semantic misalignment between tumor types despite shared histological vocabulary. 

\begin{table}[ht]
\centering
\begin{threeparttable}
\caption{\textit{Main Results: AUC-ROC Across All Experimental Conditions  }}
\label{tab:main_results_auc}
\begin{tabular}{lccc}
\toprule
\textcolor{HeaderBlue}{\textbf{Method}} &
\textcolor{HeaderBlue}{\textbf{Same-Cancer}} &
\textcolor{HeaderBlue}{\textbf{Cross-Cancer}} &
\textcolor{HeaderBlue}{\textbf{Cross-Species}} \\
\midrule
Zero-shot (Prototype)           & 64.89\%         & 56.84\%         & 63.96\%       \\
Few-shot 1\%                    & 67.67\%         & 58.02\%         & 63.50\%* \\
Few-shot 3\%                    & 68.65\%         & 59.41\%         & 63.50\%* \\
Few-shot 5\%                    & 68.99\%         & 61.58\%         & 63.80\%* \\
Few-shot 10\%                   & 71.74\%         & 65.29\%         & 63.20\%* \\
Few-shot 15\%                   & 72.49\%         & 64.81\%         & 63.60\%* \\
Few-shot 20\%                   & 72.56\%         & 66.31\%         & 63.40\%* \\
Text Anchoring (CLIP)           & 78.39\%         & 53.73\%         & 78.39\%         \\
Text Anchoring (Qwen)           & 77.80\%         & 53.95\%         & 77.80\%         \\
H-optimus-0                     & $84.97 \pm 0.39\%$ & ---          & $79.63 \pm 2.04\%$ \\
Phikon-v2                       & 84.87\% (SD n/a) & ---            & ---             \\
\bottomrule
\end{tabular}
\begin{tablenotes}
    \small
    \item \textit{Note.} Few-shot learning was not applied in the cross-species setting. These “few-shot” cross-species results use human-trained probes evaluated on canine patches without adaptation; they do not reflect in-domain fine-tuning. Standard deviations are omitted where single deterministic evaluations were performed due to fixed prototype construction.
    \item{*} Asterisks denote frozen prototype inference.
    
\end{tablenotes}
\end{threeparttable}
\end{table}

\subsection{Same-Cancer Transfer}
We next analyzed performance under same-cancer transfer, where training and evaluation were performed on the same cancer type and species. Model performance increased monotonically with the number of labeled training patches, as shown in Table 4. Even limited supervision led to measurable improvements over zero-shot inference.  

\begin{table}[ht]
\centering
\caption{Same-Cancer Performance Across Training Sizes}
\label{tab:same_cancer_training_size}
\begin{tabular}{lcccccc}
\toprule
\textcolor{HeaderBlue}{\textbf{Method}} &
\textcolor{HeaderBlue}{\textbf{Training Samples}} &
\textcolor{HeaderBlue}{\textbf{AUC-ROC}} &
\textcolor{HeaderBlue}{\textbf{Precision}} &
\textcolor{HeaderBlue}{\textbf{Recall}} &
\textcolor{HeaderBlue}{\textbf{F1-Score}} \\
\midrule
Zero-shot     & 0   & 64.89\% & 73.83\% & 62.50\% & 67.69\% \\
Few-shot 1\%  & 20  & 67.67\% & 75.98\% & 62.42\% & 68.53\% \\
Few-shot 3\%  & 61  & 68.65\% & 74.81\% & 71.00\% & 72.85\% \\
Few-shot 5\%  & 102 & 68.99\% & 75.30\% & 70.34\% & 72.73\% \\
Few-shot 10\% & 204 & 71.74\% & 76.49\% & 75.00\% & 75.74\% \\
Few-shot 15\% & 307 & 72.49\% & 76.67\% & 76.85\% & 76.76\% \\
Few-shot 20\% & 409 & 72.56\% & 76.83\% & 76.81\% & 76.82\% \\
\bottomrule
\end{tabular}
\end{table}

However, Figure 1 explains that performance gains began to saturate at higher training fractions, suggesting diminishing returns with additional supervision in the in-domain setting. 

 \begin{figure}[H]
    \centering
    \caption{\textit{Same-Cancer Performance Scaling Across Training Sizes.}}
    \label{fig:fig1}
    \includegraphics[width=0.8\textwidth]{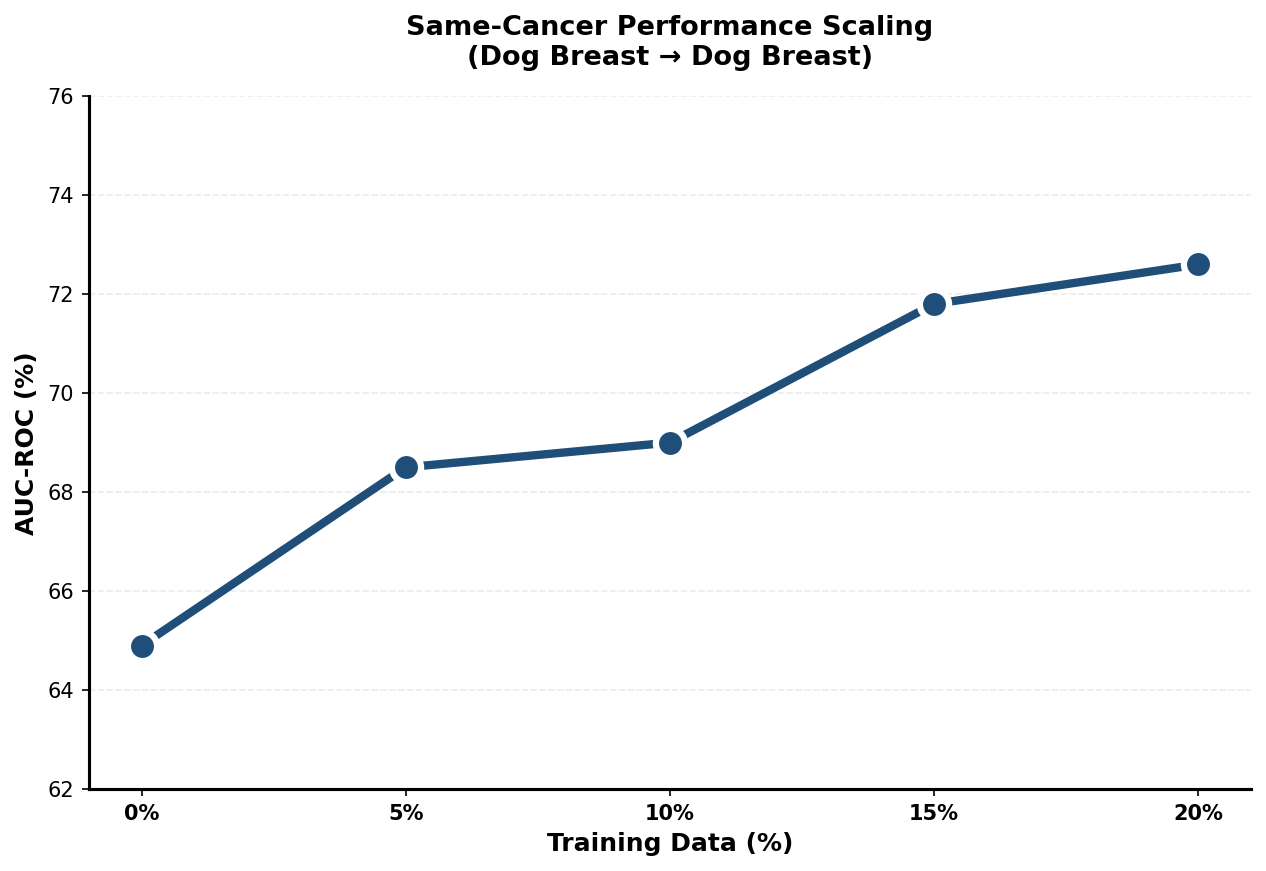}
    \par\vspace{6pt}
    \raggedright
    \small \textit{Note.} This figure illustrates the monotonic increase in AUC-ROC as the percentage of labeled training patches increases from 1\% to 20\%.
\end{figure}

 \subsection{Cross-Cancer Transfer}
 We evaluated cross-cancer transfer by training on canine breast carcinoma and testing on canine mast cell tumors. Table 5 demonstrates that this experiment showed that few-shot fine-tuning consistently improved cross-cancer performance compared to zero-shot inference. Moreover, a separate reverse-transfer experiment (human to dog) was conducted and reported values of below random (40.4\%).

 \begin{table}[ht]
\centering
\begin{threeparttable}
\caption{\textit{Cross-Cancer Performance Across Training Sizes}}
\label{tab:cross_cancer_training_size}
\begin{tabular}{lcccccc}
\toprule
\textcolor{HeaderBlue}{\textbf{Method}} &
\textcolor{HeaderBlue}{\textbf{Training Samples}} &
\textcolor{HeaderBlue}{\textbf{AUC-ROC}} &
\textcolor{HeaderBlue}{\textbf{Precision}} &
\textcolor{HeaderBlue}{\textbf{Recall}} &
\textcolor{HeaderBlue}{\textbf{F1-Score}} \\
\midrule
Zero-shot     & 0   & 56.84\% & 61.29\% & 71.68\% & 66.08\% \\
Few-shot 1\%  & 20  & 58.02\% & 65.30\% & 50.52\% & 56.97\% \\
Few-shot 3\%  & 61  & 59.41\% & 66.20\% & 49.98\% & 56.96\% \\
Few-shot 5\%  & 102 & 61.58\% & 66.80\% & 62.78\% & 64.73\% \\
Few-shot 10\% & 204 & 65.29\% & 69.39\% & 62.68\% & 65.86\% \\
Few-shot 15\% & 307 & 64.81\% & 68.89\% & 61.81\% & 65.16\% \\
Few-shot 20\% & 409 & 66.31\% & 68.62\% & 68.83\% & 68.72\% \\
\bottomrule
\end{tabular}
\begin{tablenotes}
    \small
    \item \textit{Note.} Model trained on canine breast carcinoma and evaluated on canine mast cell tumors (MITOS\_WSI\_CCMCT). 
\end{tablenotes}
\end{threeparttable}
\end{table}

Notably, training a linear classifier using human breast cancer data degrades performance substantially, producing worse-than-random results on canine mast cell tumors. Fine-tuning improved same-cancer performance monotonically but offered no benefit in cross-species settings, even at early epochs. 

\subsection{Cross-Species Transfer}

We next examined cross-species transfer by evaluating models trained on human pathology data on canine breast cancer samples. We observed that evaluating human-trained models on canine tissue raised the baseline performance to 63.96\% AUC. However, this remains below the best few-shot same-species performance (72\%), confirming that a species gap persists. Table 6 reveals that standard fine-tuning with a linear probe failed to close this gap. For this reason, we decided to see if any improvements would be made with semantic anchoring. Our goal is not to outperform H-optimus-0, but to diagnose why a vision–language pathology model underperforms despite comparable visual capacity.

\begin{table}[ht]
\centering
\begin{threeparttable}
\caption{\textit{Cross-Species Performance from TCGA-BRCA to Canine Carcinomas}}
\label{tab:cross_species_results}
\begin{tabular}{lcc}
\toprule
\textcolor{HeaderBlue}{\textbf{Method}} &
\textcolor{HeaderBlue}{\textbf{AUC-ROC}} &
\textcolor{HeaderBlue}{\textbf{Improvement}} \\
\midrule
CPath-CLIP (Prototype)        & $63.96 \pm 0.59$ (63.37–64.55) & Baseline \\
CPath-CLIP (Linear Probe)     & $71.04 \pm 1.06$ (range n/a)   & +7.08\% \\
Text Anchoring (Generic)      & $77.83$ (SD n/a)               & +13.87\% \\
Text Anchoring (Histological) & $78.28 \pm 0.50$ (77.89–78.89) & +14.32\% \\
Text Anchoring (SNOMED CT)    & $77.35$ (SD n/a)               & +13.39\% \\
Text Anchoring (Human-Specific)& $72.33$ (SD n/a)              & +8.37\% \\
Text Anchoring (Canine-Specific)& $64.79$ (SD n/a)             & +0.83\% \\
H-optimus-0 (Prototype)       & $79.63 \pm 2.04$ (range n/a)   & +15.67\% \\
\bottomrule
\end{tabular}
\begin{tablenotes}
    \small
    \item \textit{Note.} Metrics evaluate the transfer of models trained on human breast cancer (TCGA-BRCA) to canine mammary carcinoma patches. Improvement is calculated relative to the CPath-CLIP (Prototype) baseline. Standard deviations are omitted where single deterministic evaluations were performed due to fixed prototype construction
\end{tablenotes}
\end{threeparttable}
\end{table}

As shown by Figure 2, semantic anchoring substantially improves cross-species performance, achieving 78.39\% AUC, a 14.43\% improvement over the baseline and bringing CPath-CLIP within striking distance of H-optimus-0 (79.63\% AUC).

\begin{figure}[H]
    \centering
    \caption{\textit{Cross-Species Comparison of Model Performance}}    
    \label{fig:fig2}
    \includegraphics[width=0.8\textwidth]{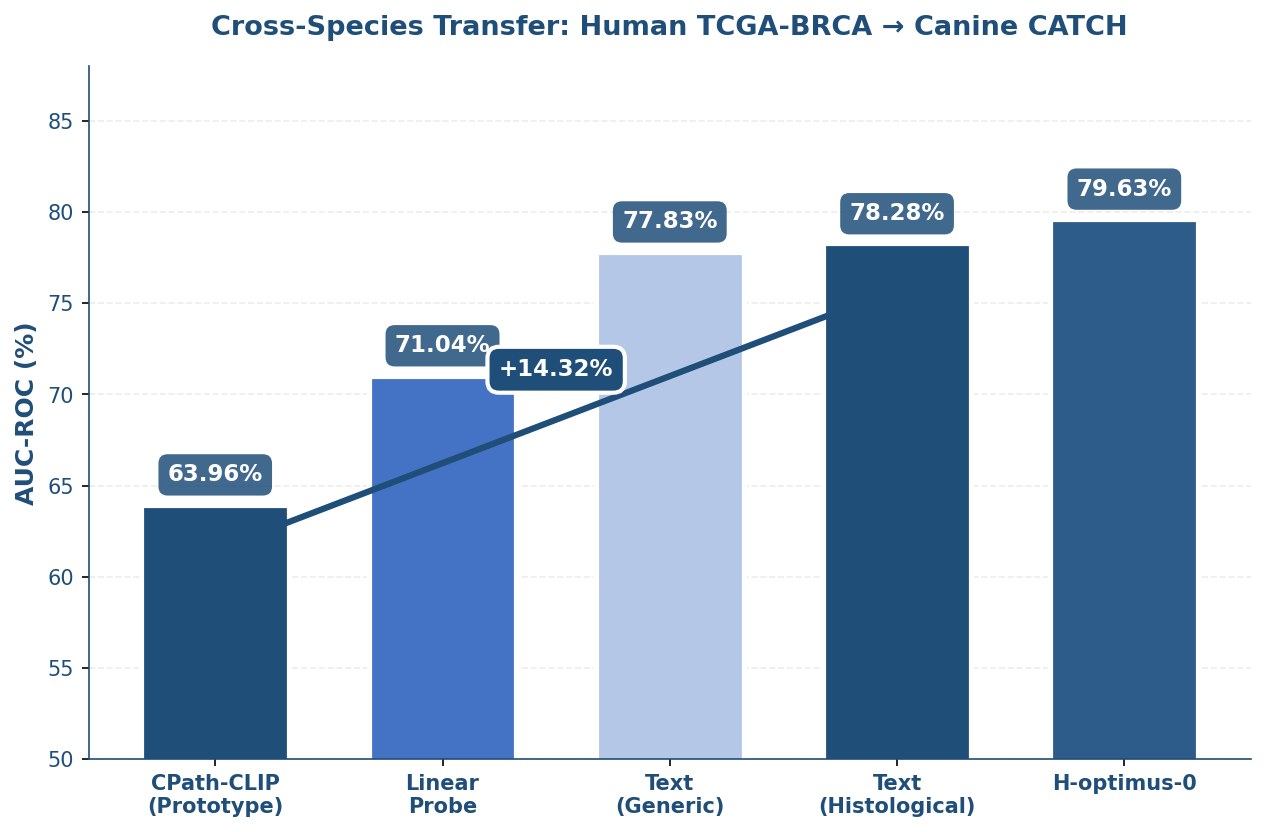}
    \par\vspace{6pt}
    \raggedright
    \small \textit{Note.} This figure compares the AUC-ROC of CPath-CLIP (zero-shot and text-anchored) against the H-optimus-0 benchmark. Semantic anchoring via language-guided alignment is shown to significantly bridge the performance gap caused by species-specific embedding collapse.
\end{figure}

Semantic Anchoring recovered cross-species performance, improving zero-shot AUC from 63.96\% to 77.80\% with Qwen2-1.5B. Linear probing with human labels partially bridges the species gap by learning a new decision boundary, but it cannot resolve the underlying semantic misalignment that text anchoring corrects. Crucially, the standard CLIP text encoder achieved nearly identical performance (78.39\%), confirming that improvement stems from the alignment mechanism itself rather than LLM sophistication. 

To determine if this improvement was driven by the sophistication of the LLM or simply by the alignment mechanism, we performed an ablation using the standard CLIP text encoder. The CLIP encoder achieved 78.39\% AUC, closely matching the LLM's performance. This confirms that the recovery of cross-species generalization is driven by the provision of a stable textual coordinate system to resolve embedding collapse, rather than the complexity of medical semantics.

Crucially, prompt choice dramatically impacts performance. Species-specific prompts (e.g., ‘Canine mammary carcinoma’) achieved only 64.8\% AUC, which was worse than generic prompts (77.8\%) and histological descriptors (78.3\%). We found that species-specific prompts suffer from high intra-prompt similarity (cosine = 0.83) because shared terms like ‘canine’ and ‘mammary’ dominate the text embedding, collapsing the semantic space. In contrast, histological prompts achieved near-maximal separation (cosine = –0.81), allowing for precise discrimination between tumors and normal tissues. This confirms that language must be engineered to avoid domain bias, not merely describe the sample. Avoid domain-identifying tokens in prompts; instead, use conserved histological descriptors.

\subsection{Embedding Space Analysis}
 
Because the foundation model's backbone was kept frozen throughout all experiments, this analysis reflects properties of the pre-trained embedding space rather than artifacts introduced by fine-tuning. Accounting for this, we analyzed the embedding space learned by the frozen foundation model to investigate the source of limited transfer performance. Tumor-vs-normal prototype cosine similarity was extremely high in some domains (e.g., dog breast prototype similarity = 0.9984), indicating severe class compression in the CLIP projection; other domains show slightly lower but still elevated prototype similarity, indicating poor intrinsic class separation. To determine if this is a general failure of pathology foundation models or specific to CPath-CLIP, we evaluated H-optimus-0 on the same 22,239 patches. H-optimus-0 achieved a prototype zero-shot AUC of 84.97\% ± 0.25\%, compared to CPath-CLIP’s 71.20\%. 

This control experiment confirms two critical findings: (1) The visual features required for cross-species detection are present in the slides, and (2) CPath-CLIP’s exhibits intrinsic embedding collapse. Notably, our text-anchoring method effectively rescues CPath-CLIP (78.28\%), bringing it within striking distance of H-optimus-0 without requiring the computationally expensive retraining of the visual backbone. 

    Figure 3 shows that cross-domain comparisons show substantially lower similarity between canine and human tumor embeddings than between canine tumor types. Moreover, Table 7 illustrates the correlation coefficient and the p-value relationships for feature importance which are moderately weak. 
\begin{figure}[H]
    \centering
    \caption{\textit{Cosine Similarity Heatmap Across Domains}}
        \label{fig:fig3}
    \includegraphics[width=0.8\textwidth]{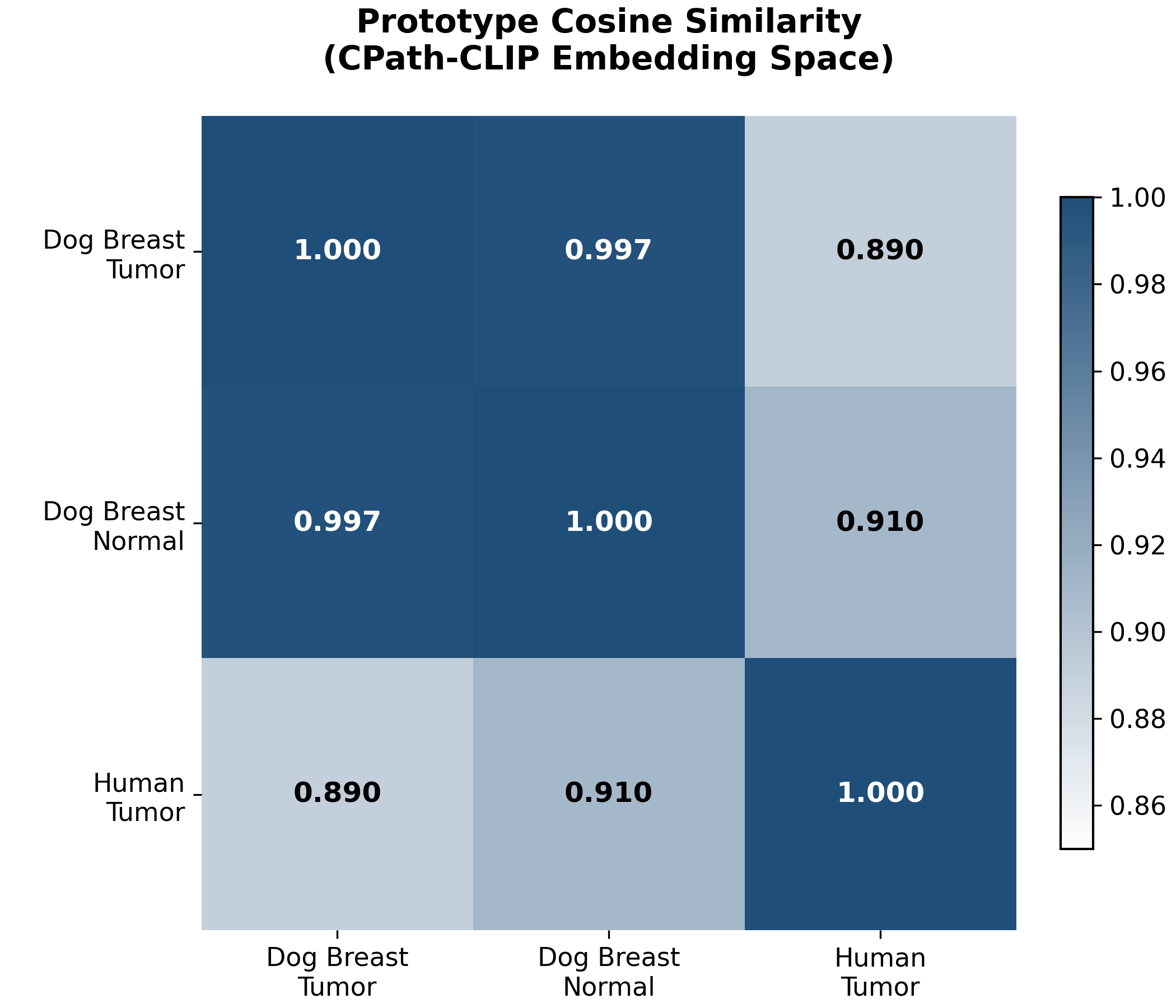}
    \par\vspace{6pt}
    \raggedright
    \small \textit{Note.} This confusion matrix visualizes the similarity between latent representations. High similarity values between tumor and normal prototypes within the same species indicate potential embedding collapse. 
\end{figure}

    \begin{table}[ht]
\centering
\begin{threeparttable}
\caption{\textit{Feature Importance Correlation Across Domains}}
\label{tab:feature_correlation}
\begin{tabular}{lccc}
\toprule
\textcolor{HeaderBlue}{\textbf{Domain Pair}} &
\textcolor{HeaderBlue}{\textbf{Pearson r}} &
\textcolor{HeaderBlue}{\textbf{p-value}} &
\textcolor{HeaderBlue}{\textbf{Interpretation}} \\
\midrule
Dog Breast $\leftrightarrow$ Dog Mast Cell & 0.387 & $<0.001$ & Weak correlation \\
Dog Breast $\leftrightarrow$ Human (TCGA)  & 0.389 & $<0.001$ & Weak correlation \\
Dog Mast Cell $\leftrightarrow$ Human (TCGA) & 0.457 & $<0.001$ & Moderate-weak \\
\bottomrule
\end{tabular}
\begin{tablenotes}
    \small
    \item \textit{Note.} Feature importance correlations are consistently weak across domains, suggesting that morphological features driving discrimination in one species or cancer type do not generalize directly to others without realignment. Weak correlations (r < 0.46) indicate poor transferability of feature importance across domains
\end{tablenotes}
\end{threeparttable}
\end{table}

 This suggests that features driving discrimination in one domain do not generalize well to others.

\subsection{Visual Attribution Analysis with Grad-CAM}
To qualitatively assess whether the model attends to biologically meaningful regions, we applied Grad-CAM to the final transformer blocks for representative tumor and normal patches across species. Grad-CAM heatmaps (Figures 4-6) revealed diffuse and overlapping activation patterns for tumor and normal tissue, consistent with the high cosine similarity observed in the embedding space.  Together, these qualitative results support the quantitative findings that the learned representations lack discriminative, species-invariant tumor features. 

\begin{figure}[H]
    
    \centering
    \caption{\textit{Grad-CAM Visualization: Canine Breast Carcinoma}}
        \label{fig:fig4}
    \includegraphics[width=0.8\textwidth]{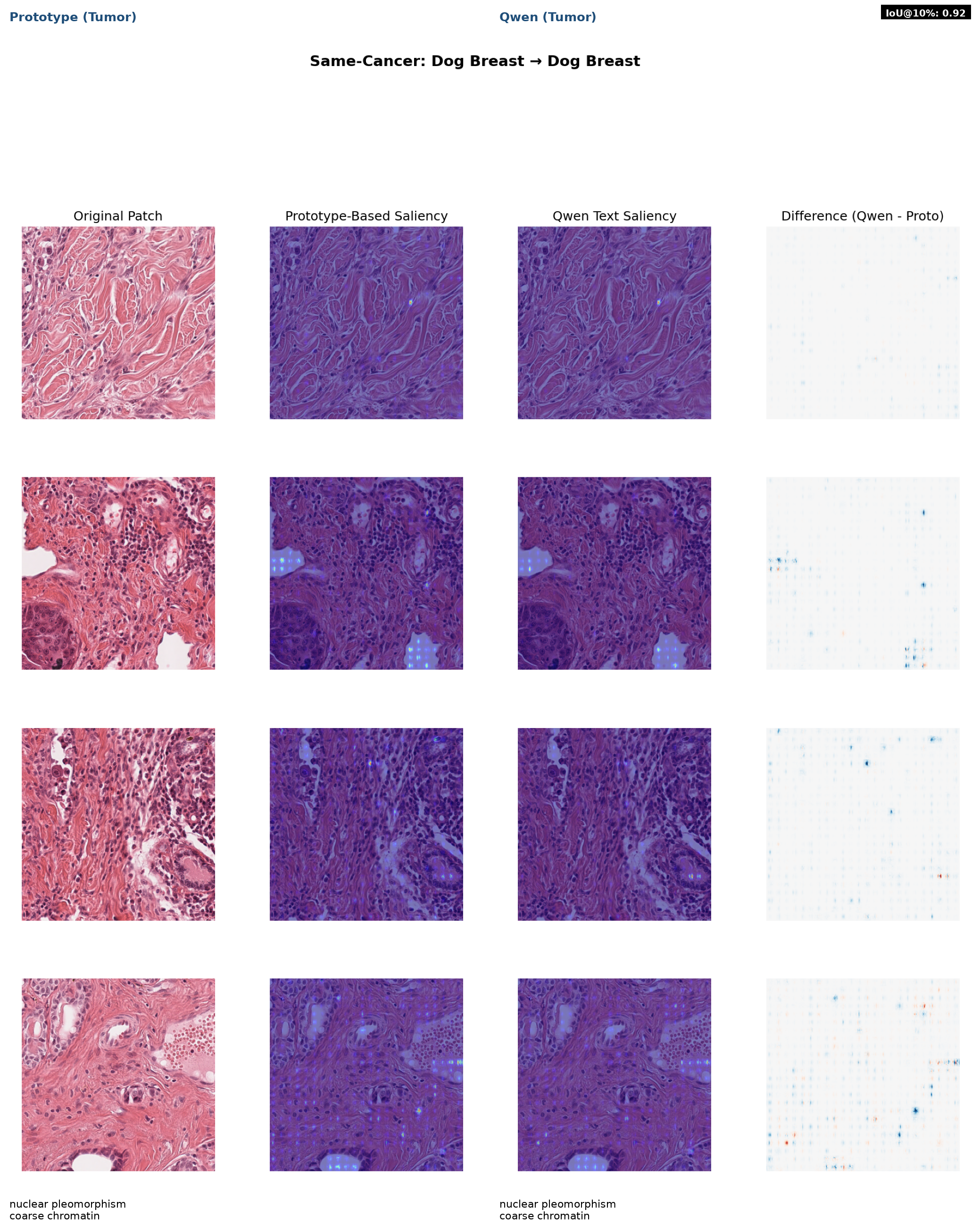}
    \par\vspace{6pt}
    \raggedright
    \small \textit{Note.} Heatmaps indicate regions of high visual activation. The diffuse pattern suggests the model is attending to broad stromal features rather than localized nuclear atypia.
\end{figure}

\begin{figure}[H]
    \centering
        \caption{\textit{Grad-CAM Visualization: Canine Mast Cell Tumor}}
            \label{fig:fig5}
    \includegraphics[width=0.8\textwidth]{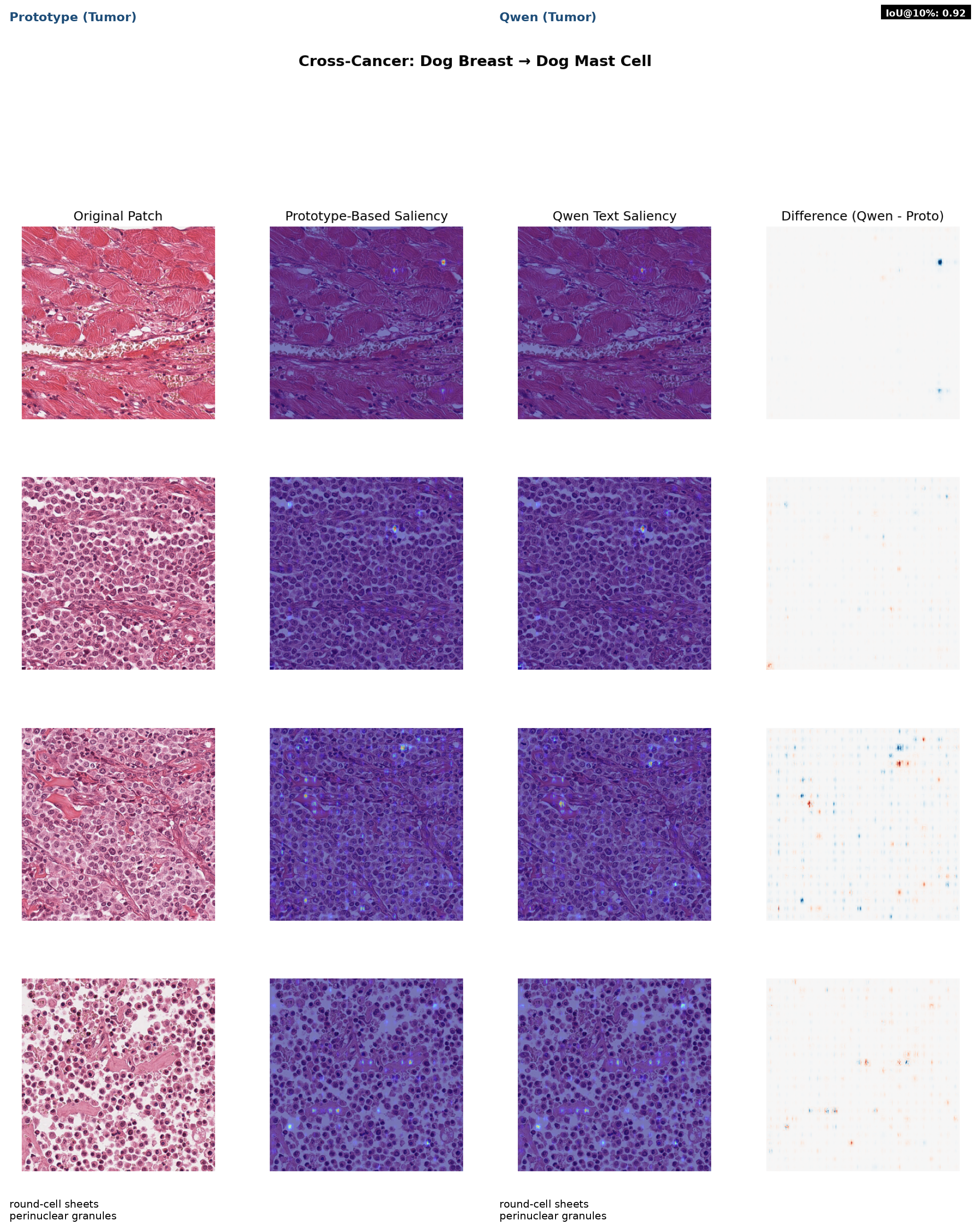}
    \par\vspace{6pt}
    \raggedright
    \small \textit
    {Note.} Visual attribution for mast cell tumors shows overlapping activation between neoplastic cells and inflammatory background, reflecting the high cosine similarity in the embedding space.
\end{figure}

\begin{figure}[H]
    \centering
        \caption{\textit{Grad-CAM Visualization: Human Breast Carcinoma to Dog Breast Carcinoma}}
    \includegraphics[width=0.8\textwidth]{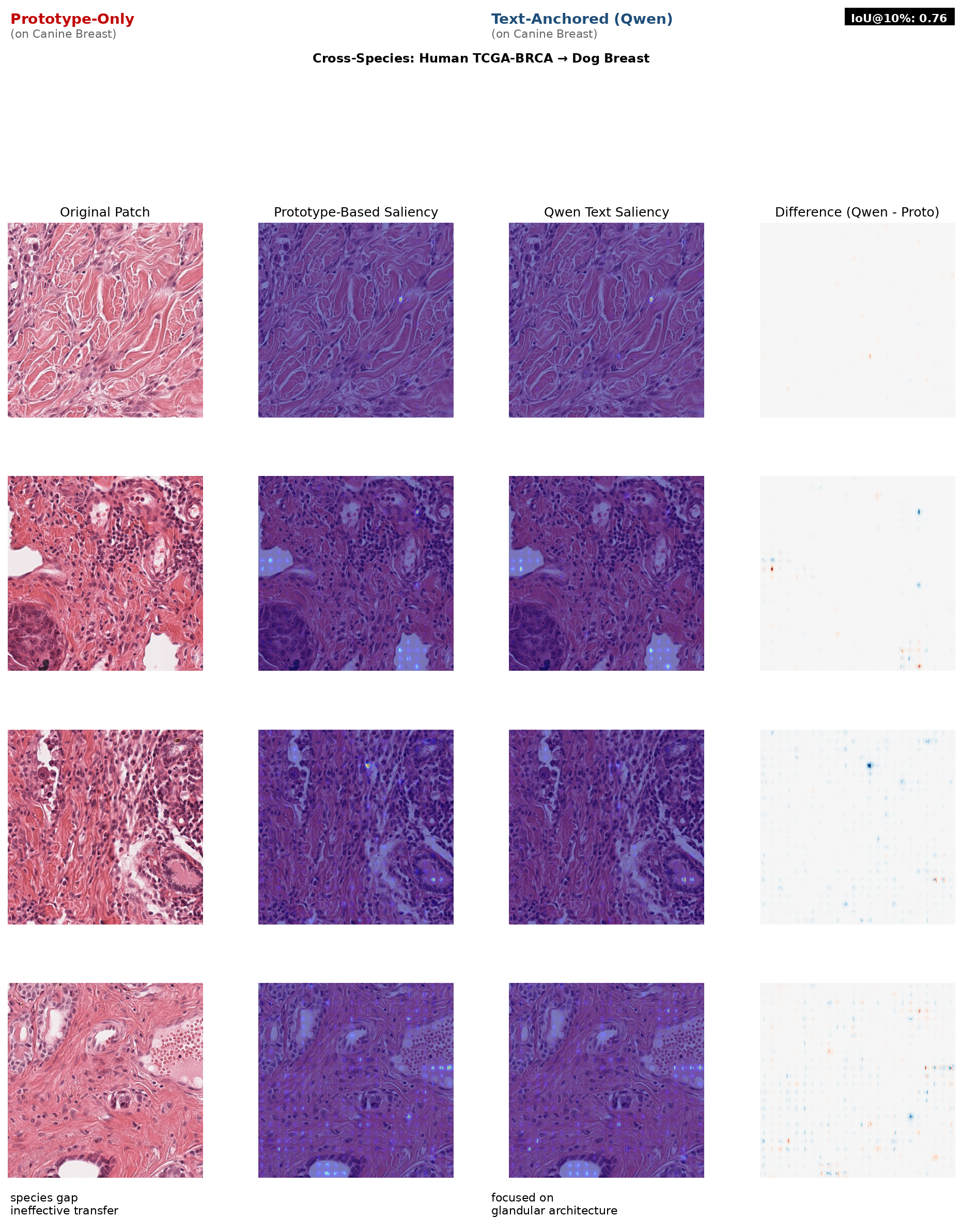}
    \label{fig:fig6}
    \par\vspace{6pt}
    \raggedright
    \small \textit{Note.} The prototype-only approach fails on canine tissue (diffuse activation), while text-anchored inference overcomes the species domain gap by grounding attention on conserved morphological features rather than species-specific visual artifacts.
\end{figure}

 \section{Discussion}
 In this study, we systematically evaluated how fine-tuning a pathology foundation model affects performance under same-cancer, cross-cancer, and cross-species transfer conditions. Our results demonstrate that few-shot fine-tuning provides consistent benefits for same-cancer and cross-cancer detection but fails to enable meaningful cross-species generalization. These findings highlight both the strengths and fundamental limitations of current pathology foundation models when deployed beyond their training domain. 

It is important to reconcile our findings of embedding collapse with the reported state-of-the-art performance of CPath-CLIP in the literature. Previous benchmarks largely rely on linear probing, a technique that trains a new classification head on top of frozen features. This effectively bypasses the pre-trained projection head, masking the underlying alignment issues. Our study, however, evaluates true zero-shot transfer, which relies entirely on the pre-trained text-image alignment. The fact that CPath-CLIP excels in linear probing (as confirmed by our own few-shot baselines) but fails in zero-shot transfer (>0.99 cosine similarity) isolates the failure mode specifically to the projection manifold, not the visual representation itself. This distinction is critical: it confirms that the model 'knows' the pathology but cannot 'speak' it without the corrected semantic coordinate system provided by our method, semantic anchoring. These findings are consistent with findings of CLIP models found by Liang. et al. 

 \subsection{The Limits of Fine-Tuning}
 
 \textbf{Fine-Tuning Improves Performance Within Species but Not Across Species} 

	Few-shot fine-tuning consistently improved tumor classification performance when the source data belonged to the same species. In the same cancer setting, fine-tuning led to consistent gains over its zero-shot counterpart, indicating that even limited task-specific supervision allowed the model to better align its representations with the target decision boundary (see Table 4).  

The successful transfer from canine breast to canine mast cell tumors is a significant finding. It demonstrates that pathology foundation models, when fine-tuned, do not simply overfit the source organ. Instead, they leverage supervision to refine a shared representation of malignancy that holds true across different cancer types. This validates the utility of CPath-CLIP for veterinary applications: if the species is consistent, the model can generalize unseen cancer types with minimal data.  

	In contrast, performance under cross-species transfer remained moderate but substantially below SOTA regardless of whether fine-tuning was applied. This failure persisted across all training regimes, indicating that few-shot supervision alone is insufficient to bridge the domain gap between species. Our benchmarking with H-optimus-0 on canine breast carcinoma patches (84.97\% AUC) indicates that the failure was not due to a lack of visual features in the slides. Instead, it stems from 'Embedding Collapse' in CPath-CLIP, where tumor and normal representations overlap almost entirely (>0.99 cosine similarity). Few-shot fine-tuning failed because the underlying topological space lacked the necessary separation to learn a valid decision boundary with limited data. 

	The observed gains from fine-tuning within species suggest that the pathology foundation model already encodes a rich set of morphologic features relevant to tumor discrimination, but that these features are not optimally organized for downstream classification without task-specific adaptation. Few-shot fine-tuning appears to act primarily as a mechanism for refining decision boundaries rather than learning entirely new visual concepts. 

	Importantly, the success of fine-tuning within species indicates that tumor-related features learned from one cancer type can remain transferable to others while the species remains constant. This finding challenges the common assumption that fine-tuning can be beneficial when the underlying representation space already captures the dominant sources of variation relevant to the task. 

	Taken together, these results imply that species-level morphology constitutes a stronger organizing principle in the embedding space than cancer type. As long as this species-level structure is preserved, fine-tuning can improve performance even when cancer-specific visual patterns differ.  
 
\subsection{Species-Dominated Semantic Collapse}

Cross-species generalization remained limited despite cross-cancer improvements. Tissue-matched transfer (TCGA-BRCA) achieved only 63.96\% AUC, far below H-optimus-0's 79.63\%,with tumor-normal embeddings exhibiting near-complete overlap (cosine similarity >0.99). This suggests 'clustering by species' suppresses tumor-discriminative features. However, the successful retrieval of performance using Qwen indicates that the visual backbone does capture species-invariant morphology; it merely lacks the projection mechanics to separate them. Our comparison with H-optimus-0 reveals that the limitation is specific to the alignment quality of CPath-CLIP, not the visual capacity of foundation models in general. The success of the standard CLIP encoder (matching Qwen) clarifies the mechanism: the text encoder acts as an orthogonal anchor, allowing the projection head to realign the collapsed visual embeddings along meaningful axes. This yields a substantial +14.43\% AUC improvement (63.96\% to 78.39\%), demonstrating that the visual backbone was already capable of capturing relevant tumor morphology but simply required more contextual knowledge to recognize it. This reveals that cross-species transfer is not limited by what the model sees, but by how it interprets what it sees, suggesting that richer semantic guidance can unlock latent knowledge in future pathology foundation models. Our prompt ablation reveals that semantic dominance is not inherent to the model but induced by poor prompt design. When prompts encode domain cues (e.g., species), they reinforce collapse; when they encode conserved morphology, they resolve it. 

	One of the most important findings of this work is that large performance gains can be achieved without changing the visual backbone at all. Across every experiment, the vision encoder was kept frozen, yet cross-species performance improved dramatically once the semantic interface to the model was changed. This shows that the pathology foundation model already encodes meaningful, species-invariant tumor features, but those features are not always accessible to downstream methods.  

Our exhaustive analysis of the embedding space provides a warning for future researchers: standard cosine similarity metrics are insufficient for evaluating pathology foundation models. The fact that fine-tuning improved performance despite a 'collapsed' embedding space (>0.99 similarity) proves that discriminative features are present in the high-dimensional manifold, even if they are compressed in the cosine projection. 

	Introducing a text encoder did not add new visual information but rather provided an orthogonal anchor to resolve the embedding collapse. This is supported by the fact that H-optimus-0 achieved high performance without any text guidance, proving the visual signal was always present. Text alignment simply allowed CPath-CLIP to recover this signal. 

 	These results suggest that pathology vision models should not be viewed as either “having” or “lacking” the correct features. Rather, they act as high-capacity feature extractors whose usefulness depends on how those features are queried. Language provides a mechanism for controlling this interpretation, allowing the same frozen visual representation to support very different downstream behaviors without retraining. 

	However, this reliance on text introduces a new variable: prompt sensitivity. Our robust experiments showed substantial performance swings based on phrasing. Paradoxically, highly specific prompts (e.g., 'Canine mammary carcinoma') underperformed simple ones ('Tumor'). Future implementations must prioritize prompt optimization, as 'semantic dominance' can backfire if the prompt over-emphasizes domain-specific keywords. 

	Cross-domain failure in computational pathology is often assumed to result from missing or insufficient visual features. Under this assumption, poor transfer performance implies that the model has not learned the correct morphology for the target domain. Our results challenge this explanation.  

	Despite complete failure of prototype-based zero-shot transfer across human to dog transfer, the same visual embeddings support strong performance once semantic alignment is improved. This indicates that the relevant tumor information was present all along but was not being used effectively. We therefore propose a new failure model for cross-domain pathology transfer: semantic dominance by species-level morphology.  

	In this setting, embeddings are primarily organized by global tissue and species-specific structures, which overwhelm subtler tumor-related signals. This creates the appearance of representation collapse when analyzed using cosine similarity, even though discriminative tumor features remain embedded in the space. Fine-tuning does not resolve this issue because it adjusts decision boundaries without changing what visual evidence the model attends to.  

	The language-guided approach disrupts this dominance by reweighting visual features based on medically meaningful concepts. Grad-CAM analysis shows that prototype-based methods remain locked to breast-specific glandular patterns even when evaluating mast cell tumors or cross-species data. In contrast, the language-guided model adapts its attention based on the semantic definition of the task, explaining its improved performance under domain shift.

    \subsection{Interpretability and Attention}
 
    To better understand how pathology foundation models make predictions, we applied Grad-CAM to visualize all the regions of interest that the model attends during classification. Rather than serving as a post-hoc visualization, these saliency maps provide insights into the representational constraints observed in earlier experiments. 
\begin{description}[style=nextline]

\item[Same-Cancer Transfer: Shared Visual Strategy Within Domain:]

	In the same cancer setting, prototype- and text-guided Grad-CAM saliency correlations averaged the Pearson r = 0.74–0.76 across the sampled patches. Both methods consistently focused on using glandular structures and regions of nuclear atypia, hallmarks of breast cancer morphology. This high correlation indicates that the baseline prototype already captures relevant visual cues when the training and evaluation domains align, leaving limited room for semantic intervention by Qwen. This shows that whenever morphology, species, and cancer type are consistent, visual statistics alone are largely sufficient for reliable classification. 

\item[Cross-Cancer Transfer: Breaking Domain-Locked Attention] 

However, a different pattern emerged in the cross-cancer setting (dog breast → dog mast cell). Prototype-based attention remained domain-locked, meaning that it continued to emphasize glandular breast-like structures even when evaluating mast cell tumors where such features are biologically irrelevant.  

We attribute the accuracy increase to language-guided attention shifting toward dispersed cell clusters and granular cytoplasmic regions characteristic of mast cell pathology. The reduced correlation between prototype and Qwen saliency maps (r = 0.69 ± 0.10) reflects this divergence in attention strategy. 

This qualitative shift directly explains the observed +5.67\% AUC improvement: Qwen does not merely sharpen existing attention but actively reorients the model toward cancer-type–specific morphology based on semantic cues. The text prompt “mast cell tumor” functions as a weak supervisory signal, guiding the model away from inappropriate visual priors learned during training and allowing it to use its more general cancer clues on this different type of cancer. The combination of both Qwen guidance and existing cancer classification allowed it to generalize to this new type of malignant tumor. 

\item[Cross-Species Transfer: Semantic Guidance Overcomes Species Bias] 

 The most revealing insights arise in one of the cross-species settings (human TCGA-BRCA→ dog breast), where prototype-based zero-shot inference failed over other foundation models. Grad-CAM maps showed that prototype attention fixates on human-specific tissue architectures that do not translate to canine histology, explaining the close-to-random AUC (63.96\%). 

Qwen, however, shifts focus on species-invariant features such as nuclear abnormalities and tissue disorganization; concepts conserved across mammalian cancers. Although the overall saliency correlation remains moderate (r = 0.74 ± 0.15), the semantic alignment of attended regions differs substantially. 

	This mechanistic difference provides a visual explanation for the dramatic +13.84\% AUC improvement. Rather than forcing visual alignment between species, Qwen enables the model to interpret unfamiliar morphology through shared oncologic concepts, effectively bridging the species gap without modifying the visual backbone. 
\end{description}

	To conclude, these results indicate that language encoders do not introduce entirely new visual features; instead, they reshape attention within an already informative embedding space.

    \subsection{Implications for Pathology AI}
While this study provides insight into the behavior of pathology foundation models under varying transfer regimes, several limitations should be acknowledged. 

First, our analysis focused on frozen visual backbones to isolate the effects of fine-tuning and semantic alignment. While this design choice enables causal interpretation, it does not explore whether limited, targeted unfreezing of visual layers could preserve cross-species generalization while improving in-domain performance. Investigating selective or modular fine-tuning strategies may offer a compromise between adaptability and representation stability. 

 	Second, our study focuses on binary tumor/normal classification. Extending this work to multi-class subtyping or histological grading tasks may reveal additional failure modes not captured in this binary setting. This suggests that some morphologic features are genuinely cancer-type–specific and may not be recoverable through language alone. Incorporating structured ontologies or hierarchical disease representations could help models reason for relationships between cancer types beyond flat class labels. 
    
    Third, although we applied Macenko stain normalization to reduce technical variation, biological and protocol-driven differences in H\&E staining between human and veterinary labs may persist. These residual staining shifts could contribute to domain misalignment, independent of morphology

Finally, our Grad-CAM analysis, while informative, provides only a coarse approximation of model attention. Future studies could employ more granular interpretability techniques, such as concept activation vectors or patch-level attribution, to further dissect how semantic signals influence feature utilization across domains. 

 	Despite these limitations, the results of this study point toward a clear direction for future pathology foundation models: improving generalization may depend less on scaling datasets or architectures and more on how visual representations are semantically grounded. Language-guided alignment offers a promising mechanism for unlocking latent generalization already present in existing models, particularly under severe domain shift such as cross-species transfer. While these strategies are valuable, our results suggest that they may not address the core limitation behind cross-domain failure. 

The recovery of cross-species performance without additional visual training indicates that semantic alignment can unlock latent capabilities already present in existing models. This shifts the focus from what the model sees to how the model interprets what it sees. Instead of repeatedly retraining visual backbones for every new domain, future models may benefit more from mechanisms that provide stronger semantic guidance over fixed visual representations. 

 	This has important practical implications. In settings where labeled data are scarce, such as veterinary pathology, rare diseases, or underrepresented populations, semantic control offers a way to reuse existing foundation models without costly retraining. It also improves interpretability, as changes in performance can be linked directly to shifts in semantic guidance rather than opaque changes in the visual embedding space. 

	Together, these findings suggest a shift in how multimodal pathology models should be designed and evaluated. Language should not be treated as a passive labeling mechanism, but as an active component that guides how visual information is interpreted. The role of the text encoder is not simply to describe images, but to impose a medically meaningful structure on high-dimensional visual embeddings. 

 	While this study uses Qwen and standard embeddings as proof of concepts, the underlying idea is model agnostic. Any language model capable of encoding rich medical semantics could serve as a semantic controller. Future work could explore ontology-driven prompts, hierarchical disease descriptions, or task-conditioned language inputs to further improve transferability without modifying visual representations. 

 	Overall, this work shows that cross-species and cross-domain pathology transfer are not fundamentally limited by what foundation models can perceive. Instead, it is limited by how those perceptions are interpreted. By treating language as a mechanism for semantic control rather than simple supervision, we outline a path toward more robust, reusable, and interpretable pathology foundation models. 

\section{Code and Data Availability}

The implementation of Semantic Anchoring, fine-tuning scripts, and evaluation code for CPath-CLIP and H-optimus-0 benchmarks are publicly available at:
https://github.com/ekansh-arora0/cross-species-pathology.

The repository includes:

Preprocessing scripts for WSI patch extraction and Macenko normalization.

Code for zero-shot prototype computation, linear probing, and adapter fine-tuning.

Implementation of Semantic Anchoring with both CLIP and Qwen text encoders.

Grad-CAM visualization scripts and embedding space analysis tools.

Exact patch indices and train/test splits for all datasets (provided in the data/ directory).

Model weights for the Semantic Anchoring projection head, representing our novel contribution, are available for download from the repository's checkpoints/ directory. The CPath-CLIP vision encoder weights are proprietary and not publicly redistributable; the pipeline is compatible with other CLIP-style vision encoders (see repository documentation).

Public datasets used in this study are available from: Canine Mammary Carcinoma [11], MITOS\_WSI\_CCMCT [12], and TCGA-BRCA [13] via The Cancer Imaging Archive.

\section*{Funding}
This research received no external funding.

\section*{Acknowledgments}
I would like to thank Dr. Torbert from Thomas Jefferson High School for Science and Technology for serving as my faculty sponsor and providing institutional oversight for this independent research project.

\printbibliography

\end{document}